\documentclass[11pt]{article}
    
    \usepackage[final]{acl}
    
    \usepackage{times}
    \usepackage{latexsym}
    \usepackage{amsmath}
    \usepackage[T1]{fontenc}
    
    \usepackage[utf8]{inputenc}     
    
    \usepackage{microtype}
    
    \usepackage{inconsolata}
    
    \usepackage{graphicx}
    \usepackage{booktabs}
    \NewDocumentCommand{\tianshi}
    { mO{} }{\textcolor{orange}
    {\textsuperscript{\textit{tianshi}}\textsf{\textbf{\small[#1]}}}}
    
    \usepackage{ulem}
    \usepackage{tcolorbox}
    \tcbuselibrary{skins, breakable}
    \newtcolorbox[auto counter, number within=section]{promptbox}[2][]{%
        colframe=blue!75!black,
        colback=blue!10,
        coltitle=white,
        fonttitle=\small\bfseries,
        title=Prompt Template~\thetcbcounter: #2,
        breakable, 
        enhanced,
        fontupper=\ttfamily,
        #1 
    }
    \title{\textsc{DixitWorld}: Evaluating Multimodal Abductive Reasoning in Vision-Language Models with Multi-Agent Dixit Gameplay}
    
\author{Yunxiang Mo\thanks{~~Equal Contribution}, Tianshi Zheng\footnotemark[1], Qing Zong, Jiayu Liu, Baixuan Xu\\ \textbf{Yauwai Yim, Chunkit Chan, Jiaxin Bai, Yangqiu Song} \\
  Department of Computer Science and Engineering, HKUST, Hong Kong SAR, China\\
  \texttt{\{ymoaj, tzhengad\}@connect.ust.hk, yqsong@cse.ust.hk}\\
}
    
\begin{document}
    \maketitle

\begin{abstract}
Multimodal abductive reasoning---the generation and selection of explanatory hypotheses from partial observations---is a cornerstone of intelligence. Current evaluations of this ability in vision--language models (VLMs) are largely confined to static, single-agent tasks. Inspired by \textit{Dixit}, we introduce \textsc{DixitWorld}\footnote{\href{https://github.com/HKUST-KnowComp/DixitWorld}{https://github.com/HKUST-KnowComp/DixitWorld}}, a comprehensive evaluation suite designed to deconstruct this challenge. \textsc{DixitWorld} features two core components: \textbf{DixitArena}, a dynamic, multi-agent environment that evaluates both hypothesis generation (a ``storyteller'' crafting cryptic clues) and hypothesis selection (``listeners'' choosing the target image from decoys) under imperfect information; and \textbf{DixitBench}, a static QA benchmark that isolates the listener's task for efficient, controlled evaluation.
Results from DixitArena reveal distinct, role-dependent behaviors: smaller open-source models often excel as creative storytellers, producing imaginative yet less discriminative clues, whereas larger proprietary models demonstrate superior overall performance, particularly as listeners. Performance on DixitBench strongly correlates with listener results in DixitArena, validating it as a reliable proxy for hypothesis selection. Our findings reveal a key trade-off between generative creativity and discriminative understanding in multimodal abductive reasoning, a central challenge for developing more balanced and capable vision-language agents.

\end{abstract}

    \begin{figure}[t]
        \centering
        \includegraphics[width=1\linewidth]{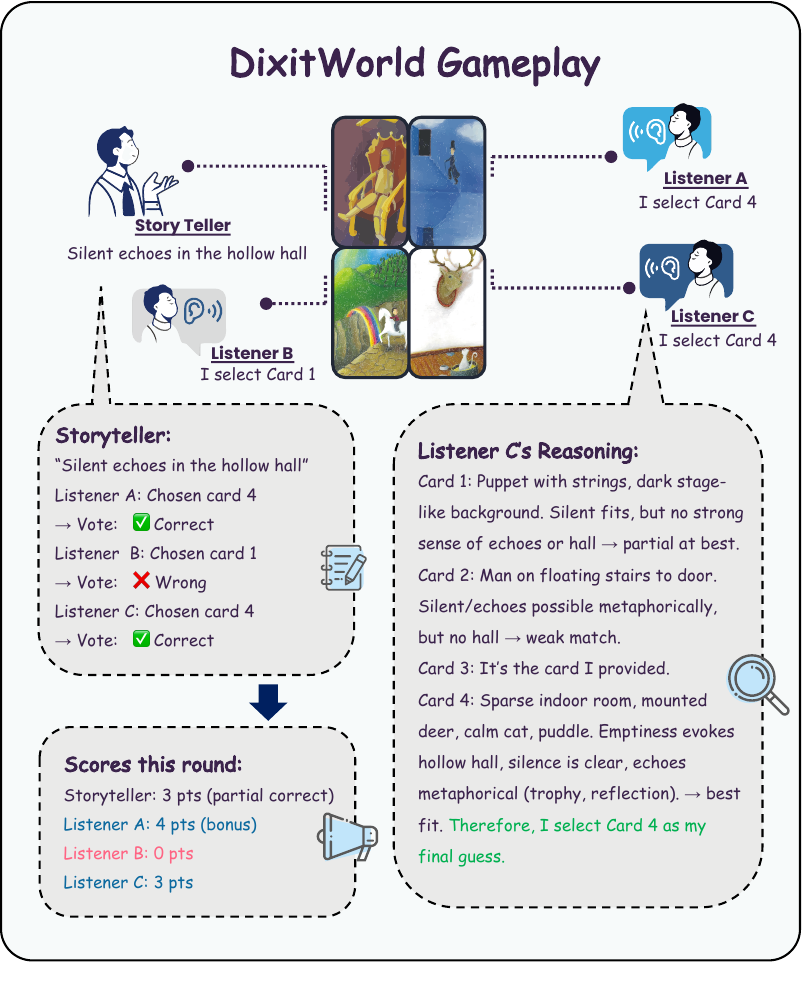}
        \caption{An illustration of Dixit gameplay.}
        \label{fig:general view}
        \vspace{-0.5cm}
    \end{figure}

  \section{Introduction}
Abductive reasoning \citep{peirce1931collected,frankfurt1958peirce} is a cornerstone of human intelligence denoting the inference process towards the best explanatory hypothesis from observation. This ability is fundamental to complex cognitive tasks ranging from commonsense reasoning to scientific discovery \citep{bhagavatula2020abductivecommonsensereasoning, bisk2019piqareasoningphysicalcommonsense}.

Recent advances in vision-language models (VLMs) \citep{liu2023visualinstructiontuning,geminiteam2025geminifamilyhighlycapable} have shown remarkable progress in perception-grounded reasoning tasks. However, the extent to which these models can perform abductive reasoning remains unclear. Prior benchmarks such as VAR \cite{liang2022visualabductivereasoning} and Sherlock \citep{hessel2022abductionsherlockholmesdataset} primarily focus on static, single-agent inference, leaving open the question of how VLMs behave in dynamic, multi-agent reasoning scenarios.

To bridge this gap, we introduce \textsc{DixitWorld}, an evaluation suite inspired by the game \textit{Dixit}. Its core component, \textbf{DixitArena}, provides a near-perfect operationalization of multimodal abductive reasoning by naturally decomposing the process into two complementary roles. The \textbf{Storyteller} performs \textbf{hypothesis generation}: given an image (the observation), they must generate a cryptic clue (the explanatory hypothesis) that is abstract enough to create ambiguity but clear enough to be solvable. Conversely, the \textbf{Listeners} perform \textbf{hypothesis selection}: presented with the clue, they must perform an "inference to the best explanation" by evaluating a set of competing visual hypotheses—the target and adversarial decoys—to identify the most plausible origin of the clue. Our large-scale simulations reveal a stark asymmetry in how models handle these roles: smaller open-source models often act as more creative but less precise storytellers, while larger proprietary models excel as listeners but struggle with the storyteller's core challenge of balancing informativeness and ambiguity.

To enable more efficient and controlled evaluation, the \textsc{DixitWorld} suite also includes \textbf{DixitBench}. This static benchmark isolates the listener's role, reframing the hypothesis selection challenge as a multiple-choice QA task with adjustable difficulty. Crucially, model performance on DixitBench strongly correlates with listener performance in DixitArena, validating it as a reliable and lightweight proxy for evaluating hypothesis selection.

Taken together, \textsc{DixitWorld} offers a fine-grained, role-based perspective on multimodal abductive reasoning. By decomposing the task into hypothesis generation and selection within a dynamic game, our analysis reveals a critical storyteller-listener performance gap, demonstrating that current VLMs excel at discrimination but falter at controlled, creative generation. This highlights the value of using interactive, multi-agent dynamics to uncover nuanced model behaviors that static benchmarks miss, paving the way for future research into developing more pragmatically sophisticated and balanced vision-language agents.

\section{\textsc{DixitWorld}}

\paragraph{DixitArena}
DixitArena is an interactive, multi-agent environment designed to operationalize multimodal abductive reasoning. In each match, four agents alternate between the roles of \textbf{Storyteller} and \textbf{Listeners}. The Storyteller performs \emph{hypothesis generation} by observing an image and creating a cryptic clue. The Listeners then perform \emph{hypothesis selection}, inferring which image best explains that clue from a set of distractors. A key feature is the scoring mechanism, which directly rewards abductive success: the Storyteller only scores points for partial correctness—when some, but not all, Listeners identify the target.

\paragraph{DixitBench}
To complement the dynamic environment, DixitBench is a static benchmark designed to isolate and control the evaluation of the listener's task. It reframes hypothesis selection as a multiple-choice QA problem, where models must identify the target image for a given clue from a set of distractors. The benchmark consists of 168 questions in total, with distractor difficulty systematically controlled via semantic similarity between pre-generated captions. This design creates distinct \textit{Easy} and \textit{Hard} subsets for fine-grained analysis. As performance on DixitBench strongly correlates with listener results in DixitArena, it serves as a validated and efficient proxy for this ability.

\paragraph{Evaluation Metrics}  
Performance in \textbf{DixitArena} is measured using three primary metrics: (i) \textbf{Storyteller Score}, the percentage of rounds achieving partial correctness; (ii) \textbf{Listener Accuracy}, the percentage of correct target identifications; and (iii) \textbf{Overall Score}, the average of the Storyteller and Listener scores. 

The overall normalized score for a model is the average of its normalized scores across all matches, calculated as:
\begin{equation}
    Score = \left( \frac{1}{| \mathcal{N} |}
    \sum_{i \in \mathcal{N}}
    \frac{\text{score}_{i}^{\text{attained}}}{\text{score}_{i}^{\text{max}}} \right) \times 100\%
\end{equation}
where $\mathcal{N}$ represents the set of matches played.

For DixitBench, we report overall accuracy as well as performance across the Easy and Hard subsets. All scores are normalized to a 0–100\% scale. Full implementation details, fairness procedures, and difficulty calibration are provided in Appendix~\ref{app:benchmark} and \ref{app:setup}.

\section{Experiment and Analysis}

We evaluate six modern VLMs of varying scales, including Qwen2.5-VL-7B/32B, Gemini-2.5-Flash, Gemma3-12B/27B, and GPT-4o. All models are configured with the temperature of 0.7 to encourage diversity. Model specifications and prompt templates are detailed in Appendix~\ref{app:models} and \ref{app:prompt}.

\subsection{DixitArena Gameplay Performance}

\begin{figure}[t]
 \centering
 \includegraphics[width=\linewidth]{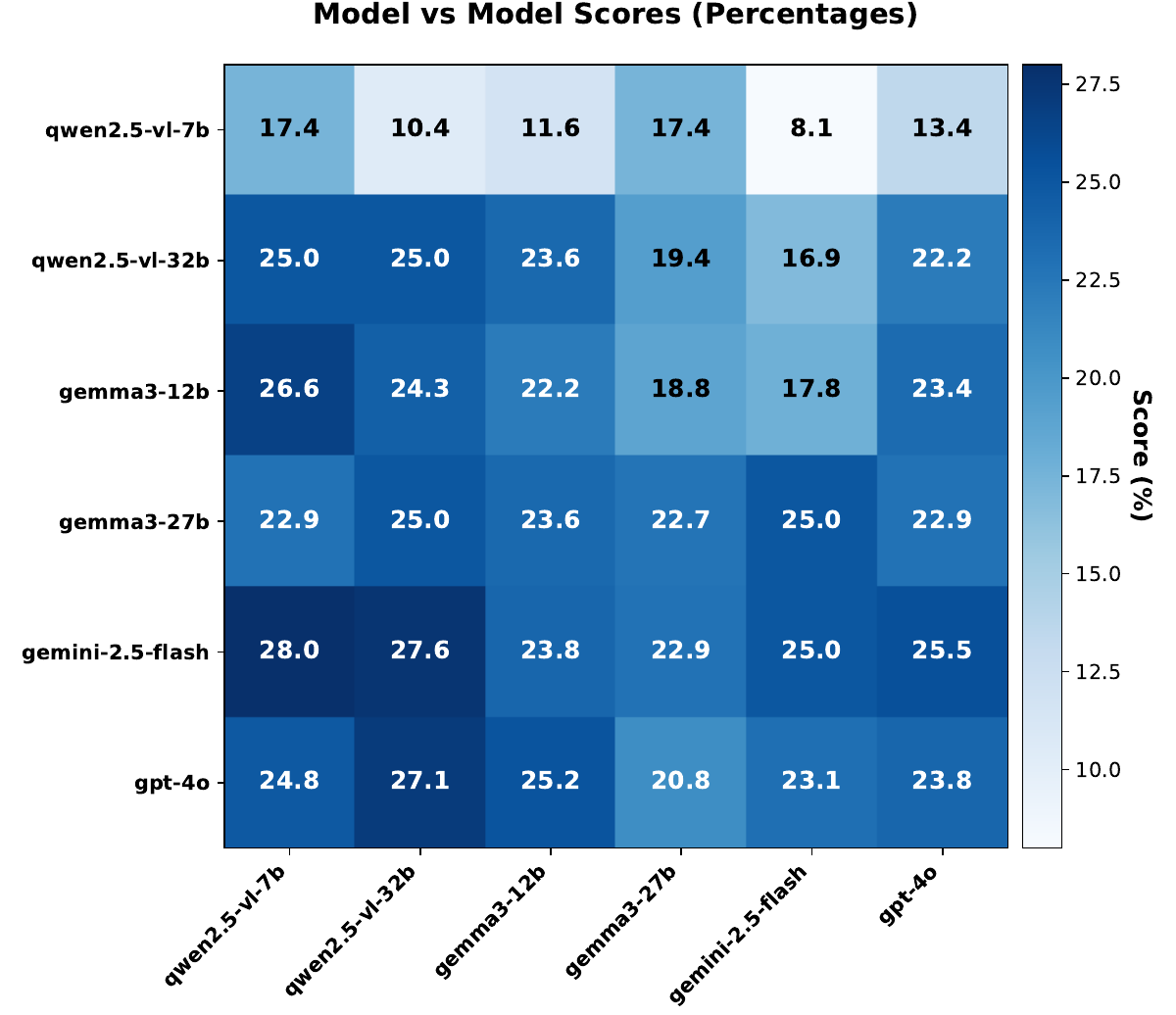}
 \caption{Head-to-head game scores for all model pairings in the round-robin tournament, demonstrating consistent performance hierarchies.}
 \label{fig:model_vs_model}
 \vspace{-0.3cm}
\end{figure}

\begin{table}[t]
\centering
\small

\begin{tabular}{lccc}
\toprule
\textbf{Model} & \textbf{Storyteller} & \textbf{Listener} & \textbf{Overall} \\
\midrule
Qwen2.5-VL-7B    & 14.29 & 30.87 & 17.39 \\
Qwen2.5-VL-32B   & 15.48 & 50.79 & 30.63 \\
Gemma3-12B       & 20.24 & 49.44 & 31.68 \\
Gemma3-27B       & \textbf{32.14} & 49.76 & 34.15 \\
GPT-4o           & 19.05 & \underline{53.89} & \underline{34.58} \\
Gemini-2.5-Flash & \underline{30.95} & \textbf{54.05} & \textbf{36.52} \\
\bottomrule
\end{tabular}
\caption{Normalized scores (\%) for Storyteller, Listener, and Overall performance. Models are ranked by their overall average score. The best and second-best scores in each column are in \textbf{bold} and \underline{underlined}, respectively.}
\label{tab:final_results}
\end{table}

Figure \ref{fig:model_vs_model} illustrated the head-to-head gameplay performances in our round-robin tournament across all six models in DixitArena.

Larger and proprietary models (\textit{Gemini-2.5-Flash}, \textit{GPT-4o}) generally outperform smaller open-source ones, confirming that abductive reasoning benefits from scale and more comprehensive multimodal pretraining. 
In addition, models’ scores are influenced by the strength of their opponents—when facing stronger models with higher average performance, weaker models’ scores tend to drop, indicating that high-performing agents impose stronger abductive pressure through clearer or more discriminative cues. 
Conversely, matches between weaker models exhibit higher variance and less stable gameplay performance. 
Overall, the matrix reveals a consistent capability hierarchy alongside non-trivial interaction effects between model strength and opponent style.

Furthermore, Table \ref{tab:final_results} present the gameplay performance with different roles isolated. Listener performance generally scales with model size and proprietary status, while storyteller ability does not. This reveals that while large models are adept at \textit{understanding} clues, they struggle with \textit{crafting} them with the requisite balance of ambiguity and clarity. For instance, GPT-4o excels as a listener yet ranks only mid-level as a storyteller, underscoring this performance gap.

\begin{figure}[t]
 \centering
 \includegraphics[width=\linewidth]{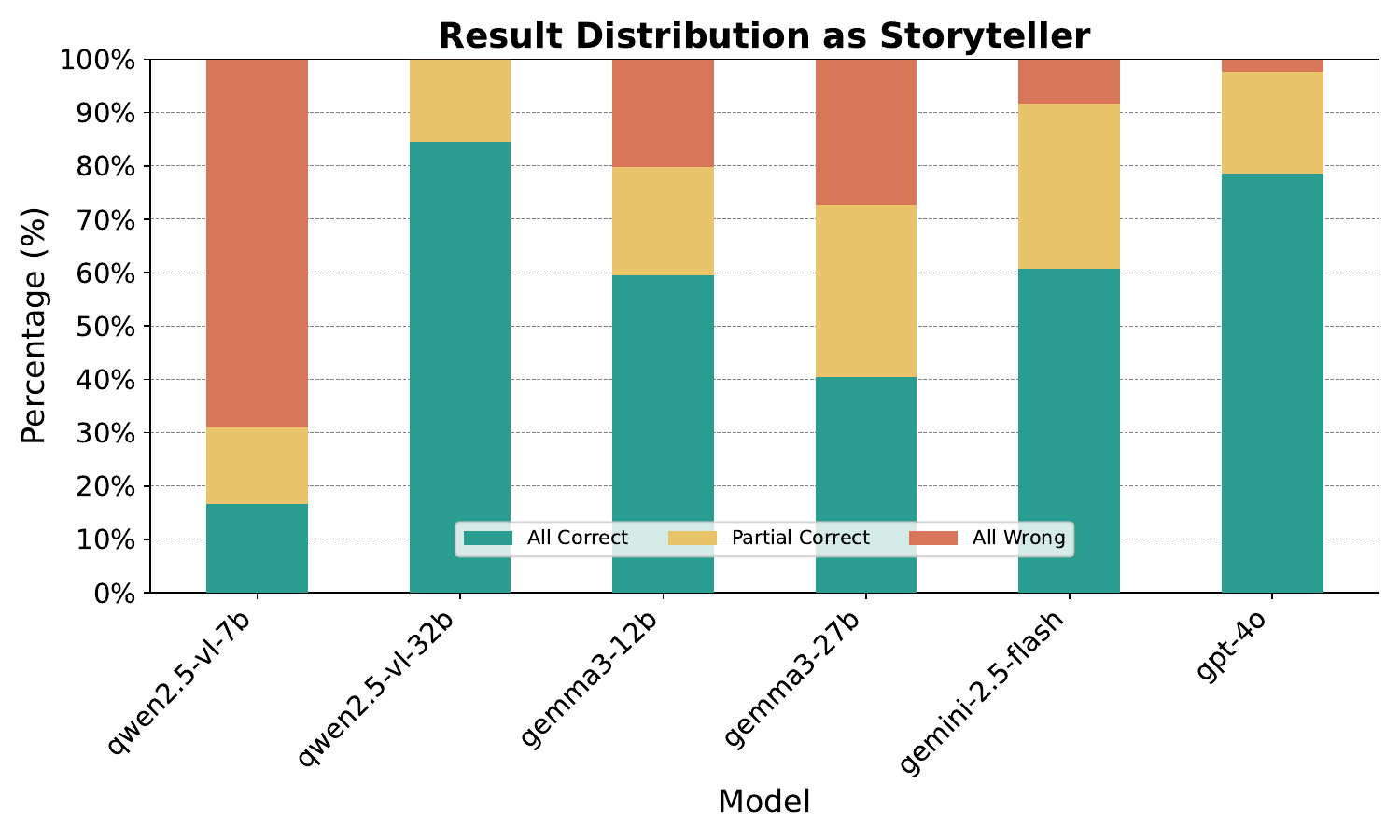}
 \caption{Distribution of storyteller round outcomes. Only the ``Partial-Correct'' outcome yields points for the storyteller, making it the desired result.}
 \label{fig:storyteller_stacked}
\end{figure}
To further diagnose the asymmetry in storyteller performance, we decomposed round outcomes into three categories: \textit{Partial-Correct} (the optimal outcome), \textit{All-Correct} (too obvious), and \textit{All-Wrong} (too vague or misleading). As shown in Figure~\ref{fig:storyteller_stacked}, over 70\% of storyteller rounds yielded zero points, falling into either the All-Correct or All-Wrong categories. This high failure rate reveals that models frequently struggle to strike the intended “Dixit balance” between clarity and ambiguity. The nature of this failure varies by model scale: smaller models often produce overly literal or vague clues, while larger ones tend toward over-specificity. This confirms that generative abduction—requiring a blend of creativity and controlled uncertainty—remains a primary bottleneck for current VLMs.

\subsection{DixitBench Performance}

\begin{table}[t]
\centering
\small
\setlength{\tabcolsep}{10pt}

\begin{tabular}{lccc}
\toprule
\textbf{Model} & \textbf{Easy} & \textbf{Hard} & \textbf{Total} \\
\midrule
Qwen2.5-VL-7B & 39.05 & 34.29 & 36.67 \\
Qwen2.5-VL-32B & 70.24 & 55.95 & 63.10 \\
Gemma3-12B & 58.33 & 57.14 & 57.74 \\
Gemma3-27B & 63.10 & \underline{61.90} & 62.50 \\
Gemini-2.5-Flash & \underline{65.48} & 64.29 & \underline{64.89} \\
GPT-4o & \textbf{78.57} & \textbf{72.62} & \textbf{75.60} \\
\bottomrule
\end{tabular}
\caption{Performance on the DixitBench. 
The small gap between Easy and Hard subsets suggests limited sensitivity to semantic difficulty.}
\label{tab:caption_similarity_results}
\vspace{-0.3cm}
\end{table}

Table~\ref{tab:caption_similarity_results} presents VLM performances on DixitBench, which are highly consistent with listener results from DixitArena, thus validating the benchmark's reliability as a proxy for hypothesis selection. The results also show that the difficulty classification based on semantic distance yields only a modest performance gap (4.8\% in average) between \textit{Easy} and \textit{Hard} questions. Although directionally valid, this modest gap shows that simple semantic distance is not a strong indicator of difficulty for the abstract gameplay scenarios in Dixit.

\begin{figure}[t]
 \centering
 \includegraphics[width=\linewidth]{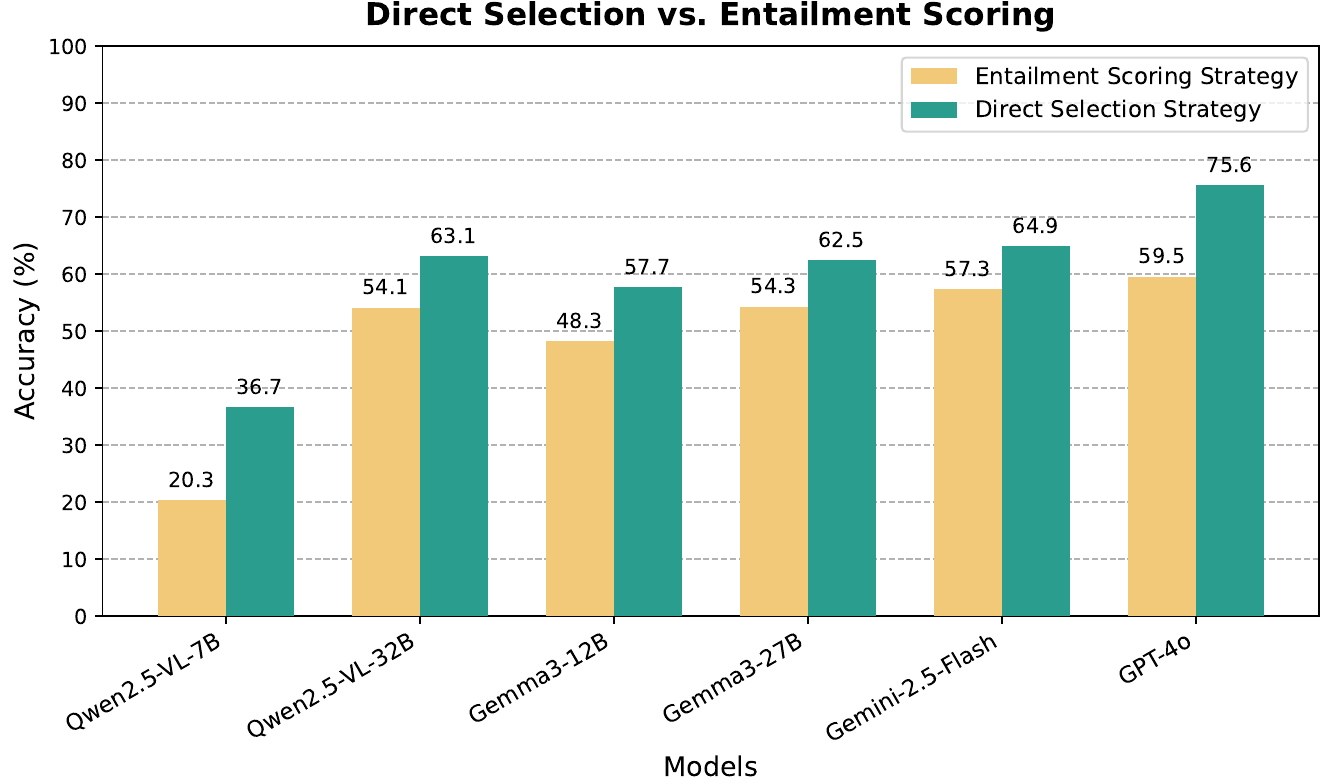}
 \caption{Comparison between direct selection and entailment scoring strategies in DixitBench.}
 \label{fig:strategy_comparison}
 \vspace{-0.2cm}
\end{figure}

We further compared listener performance under two hypothesis selection strategies: 
\textit{direct selection} (choosing one image) and \textit{entailment scoring} (rating all candidates independently and choose the highest one).  
As shown in Figure~\ref{fig:strategy_comparison}, all models perform consistently better under direct selection, 
with accuracy drops of roughly 5–20 percentage points when switching to entailment scoring 
(complete results in Appendix~\ref{app:entailment}).  
These results indicate that absolute plausibility rating introduces calibration noise and reduces model separability, 
particularly when distractors are semantically close.  
In short, VLMs are more reliable at \textit{selective abductive inference} (choosing among alternatives) 
than at \textit{discriminative entailment judgment}.

\subsection{Human Evaluation}

To complement our automated metrics, we conducted a human evaluation of the storyteller clues. We assessed the clues on two rated dimensions, \textbf{Clarity} and \textbf{Creativity}, and one performance-based dimension, \textbf{Listener Accuracy}. For the rated dimensions, human annotators scored each clue. Clarity was then calculated using our three-step robust index (detailed in Appendix \ref{app:clarity}), while creativity scores represent the normalized mean of the ratings. Listener Accuracy was measured directly as the success rate of human players using the AI-generated clues.

As demonstrated in Table \ref{tab:human_eval}, human listeners achieved 71.6\% accuracy with the AI clues, a score closely matched by Gemini-2.5-Flash (70.2\%). Among the models, a trade-off was evident: Gemma3-27B achieved the highest clarity balance (0.387), while Gemma3-12B scored highest on creativity (0.574). These results highlight a tension between informativeness and imagination, indicating that effective abductive communication demands \textit{moderate clarity with nontrivial abstraction}. This mirrors the storyteller failure patterns observed earlier, where overly explicit models lose ambiguity and overly creative ones lose alignment with the target image.

\subsection{Further Discussions}

We performed several checks to validate our evaluation's robustness and fairness. A leave-one-listener-out (LOLO) analysis confirmed stable scores across listener subsets (Appendix~\ref{app:lolo}), while a chi-squared test showed a uniform distribution of candidate image positions ($\chi^2=1.83, p=0.61$). Hand-swap phases also effectively minimized bias, with score differences remaining under 0.3 points. These results affirm the stability and impartiality of our environment.

Overall, our analysis indicates that while current VLMs possess strong discriminative grounding, they lack sophisticated pragmatic control. They succeed at hypothesis selection but fail to navigate the communicative ambiguity required for hypothesis generation. Bridging this storyteller–listener asymmetry will likely require  explicitly modeling of theory-of-mind, uncertainty, and ambiguity.

\begin{table}[t]
\centering
\small

\begin{tabular}{lccc}
\toprule
\textbf{Model} & \textbf{Acc (\%)} & \textbf{Clarity} & \textbf{Creativity} \\
\midrule
Qwen2.5-VL-7B    & 25.79 & 0.104 & 0.212 \\
Qwen2.5-VL-32B   & 66.27 & 0.188 & 0.243 \\
Gemma3-12B       & 63.49 & 0.295 & \textbf{0.574} \\
Gemma3-27B       & 65.08 & \textbf{0.387} & 0.536 \\
Gemini-2.5-Flash & 70.24 & 0.310 & 0.488 \\
GPT-4o           & 68.65 & 0.232 & 0.484 \\
\midrule
Human            & \textbf{71.59} & --- & --- \\
\bottomrule
\end{tabular}
\caption{Human evaluation results. 
``Clarity'' is computed using our three-step median–penalty formula. 
Creativity is normalized to [0,1].}
\label{tab:human_eval}
\vspace{-0.3cm}
\end{table}
\section{Conclusion}
This paper demonstrates that current VLMs, despite their strong discriminative abilities, lack the pragmatic control necessary for robust multimodal abductive reasoning. Through our novel evaluation suite, \textsc{DixitWorld}, we reveal a critical performance gap: models are proficient listeners (hypothesis selectors) but poor storytellers (hypothesis generators), struggling to balance creativity with communicative intent. This core failure to manage ambiguity highlights that discriminative success in static QA tasks does not guarantee generative competence in dynamic, goal-oriented scenarios. Advancing the field will therefore require a shift towards architectures that explicitly reason about ambiguity and model communicative intent.


 \section*{Limitations}
Our study has several limitations that future work may address:
\begin{itemize}
    \item \textbf{Task Scope.} While we operationalizes abductive reasoning through the game \textit{Dixit}, it represents only one class of multimodal abductive settings. Real-world abductive reasoning often involves temporal, causal, or interactive elements that our static and turn-based setup does not fully capture.
    \item \textbf{Model Coverage.} We evaluated a representative but limited set of modern VLMs (Qwen, Gemma, Gemini, GPT families). Broader coverage—including models fine-tuned for visual entailment or narrative generation—would further validate our conclusions about role asymmetry and entailment calibration.
    \item \textbf{Prompt and Decoding Sensitivity.} We used a standardized prompting and decoding configuration (temperature 0.7, JSON output) for comparability. Performance may vary with prompt wording, sampling parameters, or instruction tuning. A systematic study of strategy sensitivity remains future work.
    \item \textbf{Evaluation Strategy Dependence.} Our results show that performance can shift notably depending on the evaluation protocol (e.g., direct selection vs.\ entailment scoring). Future work should explore hybrid or adaptive evaluation strategies that better capture pragmatic reasoning under uncertainty.
\end{itemize}

  \section*{Ethics Statement}
This research utilizes publicly available models and visual assets. Human evaluations were conducted with compensated undergraduate students who provided informed consent and were trained on the task. No personal or sensitive data were collected, and no human participants were directly involved beyond voluntary evaluation of anonymized examples. We acknowledge that the models may reflect societal biases from their training data, which could manifest in clue generation or interpretation. Therefore, we caution against deploying automated abductive reasoning systems in sensitive contexts—such as psychological profiling or cultural analysis—without robust fairness auditing and mandatory human oversight to mitigate potential harms.

    \bibliography{main}
    
    \appendix

    \section{Related Work}
\label{app:related}

\subsection{LLMs in Multi-Agent Environments}
Recent research has increasingly explored the use of Large Language Models (LLMs) as autonomous agents capable of reasoning and planning in interactive environments \citep{zheng2023judgingllmasajudgemtbenchchatbot, openai2024gpt4technicalreport, wei2022emergentabilitieslargelanguage, yao2023treethoughtsdeliberateproblem}. When augmented with capabilities like memory, belief modeling, or tool usage, LLMs can perform complex multi-step reasoning, often outperforming traditional reinforcement learning methods in settings such as open-domain survival games and two-player imperfect information games \citep{10.1145/3586183.3606763, pmlr-v162-huang22a, guo2024suspicionagentplayingimperfectinformation,yim2024evaluatingenhancingllmsagent}. Another line of work has focused on multi-agent coordination, using both simulated environments \citep{NEURIPS2019_f5b1b89d, 10.1093/oso/9780198862536.003.0008, agashe2025llmcoordinationevaluatinganalyzingmultiagent, Li_2023} and models of human collaboration \citep{Riedl2021QuantifyingCI}. While this body of research demonstrates LLMs' potential for strategic planning, it has primarily centered on tasks with clear, instrumental objectives. In contrast, our work investigates the more nuanced challenge of creative communication, where agents must convey and interpret ambiguous concepts under imperfect information.

\subsection{Abductive Reasoning}
Abductive reasoning, the process of generating and verifying hypotheses to best explain observations \citep{peirce1931collected}, is a long-standing challenge in AI. In NLP, this ability has been predominantly studied through static, single-turn benchmarks. Early works such as aNLI \citep{bhagavatula2020abductivecommonsensereasoning} and PIQA \citep{bisk2019piqareasoningphysicalcommonsense} focused on text-based abductive inference, which was later extended to the visual domain with datasets like VisualCOMET \citep{park2020visualcometreasoningdynamiccontext}. Besides, abductive reasoning have been further applied in knowledge graph reasoning \citep{bai2024advancingabductivereasoningknowledge,gao2025controllablelogicalhypothesisgeneration}. More recent approaches have explored structured abductive inference to improve logical consistency in LLMs \citep{zheng2025logidynamicsunravelingdynamicslogical,zheng2025cursecotlimitationschainofthought,liu2025literaturemeetsdatasynergistic}, and eventually empower various downstream applications \cite{zheng2025automationautonomysurveylarge, he2025ideaenhancingrulelearning,liang2025llmhanabievaluatingmultiagentgameplays,xu2025cognitivebandwidthbottleneckshifting,zheng2025newtonbenchbenchmarkinggeneralizablescientific}. However, these single-agent paradigms fail to capture the dynamic, communicative nature of abduction as it occurs in collaborative problem-solving. Our work fills this gap through an interactive, multi-agent setting that jointly evaluates hypothesis generation and hypothesis selection in multimodal contexts.

    \section{Model Details}
    \label{app:models}
    In our experiments, we evaluated 6 modern VLMs/MLLMs with detailed information as follows:
    \begin{itemize}
        \item \textbf{Qwen2.5-VL-7B / Qwen2.5-VL-32B} \cite{bai2025qwen25vltechnicalreport} are open-source multimodal large language models developed by Alibaba, supporting both text and vision understanding with strong scaling performance across different parameter sizes.
        \item \textbf{Gemma3-12B / Gemma3-27B} \cite{gemmateam2025gemma3technicalreport} are the latest generation of Google's Gemma open-source models, designed with improved efficiency and performance for both language and multimodal reasoning tasks.
        \item \textbf{Gemini-2.5-Flash} \cite{geminiteam2025geminifamilyhighlycapable} is a proprietary multimodal model from Google, optimized for efficiency and fast inference, while maintaining competitive reasoning ability.
        \item \textbf{GPT-4o} \cite{openai2024gpt4technicalreport} is OpenAI's flagship multimodal model, supporting joint reasoning over text and images with strong listener performance in our evaluation.
    \end{itemize}

All models are accessed through the OpenRouter API, with temperature set to 0.7, with total cost of approximately \$100 USD.

    \newpage
    \section{Prompt Details}
    \label{app:prompt}
    
    \begin{promptbox}[colback=black!10, colframe=white!50!black, title=Storyteller: Select Target Image]{}
    \scriptsize
    You are a storyteller in a Dixit game. You must select one card from your 4-card hand as the target image.  
    Your goal: maximize your own score by ensuring some, but not all, players guess correctly.  
    
    IMPORTANT: Respond strictly in JSON format:  
    \begin{verbatim}
    {
        "reasoning": "Brief analysis (max 50 words)",
        "answer": "The card number (1-4)"
    }
    \end{verbatim}
    \end{promptbox}
    
    \begin{promptbox}[colback=black!10, colframe=white!50!black, title=Storyteller: Generate Description]{}
    \scriptsize
    You are a storyteller in a Dixit game. Create a description for your chosen image.  
    Scoring rules:  
    - All guess correctly = 0 points  
    - None guess correctly = 0 points  
    - Some guess correctly = 3 points (optimal)  
    
    Your description should balance ambiguity and clarity, using metaphorical and emotional language.  
    
    IMPORTANT: Respond strictly in JSON format:  
    \begin{verbatim}
    {
        "reasoning": "Reasoning towards final description",
        "answer": "Your crafted Dixit description"
    }
    \end{verbatim}
    \end{promptbox}
    
    \begin{promptbox}[colback=black!10, colframe=white!50!black, title=Listener: Select Distractor Image]{}
    \scriptsize
    You are a player in a Dixit game. Given the storyteller's description, choose one card from your hand that could plausibly match it (but is not the target).  
    Your goal: mislead others into choosing your card.  
    
    IMPORTANT: Respond strictly in JSON format:  
    \begin{verbatim}
    {
        "reasoning": "Brief analysis (max 50 words)",
        "answer": "The card number (1-4)"
    }
    \end{verbatim}
    \end{promptbox}
    
    \begin{promptbox}[colback=black!10, colframe=white!50!black, title=Listener: Direct Selection Strategy]{}
    \scriptsize
    You are a player in a Dixit game trying to guess which image matches the storyteller's description.  
    Evaluate all candidates and select the one that best fits.  
    
    IMPORTANT: Respond strictly in JSON format:  
    \begin{verbatim}
    {
        "reasoning": "Brief analysis (max 50 words)",
        "answer": "The candidate number (1-N)"
    }
    \end{verbatim}
    \end{promptbox}
    
    \begin{promptbox}[colback=black!10, colframe=white!50!black, title=Listener: Entailment Scoring Strategy]{}
    \scriptsize
    You are evaluating how well an image matches a given clue in a Dixit game.  
    Provide a numerical score (0-100) with reasoning.  
    
    IMPORTANT: Respond strictly in JSON format:  
    \begin{verbatim}
    {
        "reasoning": "Detailed reasoning for the score",
        "answer": "Your numerical rating (0-100)"
    }
    \end{verbatim}
    \end{promptbox}

\newpage
\section{Computation of the Clarity Metric}
\label{app:clarity}

For transparency, we provide the exact computation procedure for the ``Clarity'' metric used in human evaluation.  
This metric measures how well a storyteller’s clue balances ambiguity and precision, penalizing both overly vague and overly literal descriptions.  
The computation follows a three-step normalization and penalty process:

\begin{enumerate}
    \item \textbf{Linear transformation:}  
    Each raw rating $s$ (from 1 to 5) is mapped to a centered clarity score $S^\star$:
    \begin{equation}
    S^\star = 1 - \frac{|s - 3|}{2}, \quad s \in \{1,2,3,4,5\}.
    \end{equation}
    This yields $S^\star=1.0$ for $s=3$ (perfect balance), $0.5$ for $s=2,4$, and $0.0$ for $s=1,5$.

    \item \textbf{Rater aggregation:}  
    For each clue, multiple raters’ scores are aggregated by taking the median of their $S^\star$ values:
    \begin{equation}
    \text{Clarity}_{clue} = \text{median}(S^\star_1, \dots, S^\star_n).
    \end{equation}
    Using the median rather than the mean prevents misleading ``false middle'' effects (e.g., half of raters giving 1 and half giving 5).

    \item \textbf{Extreme penalty:}  
    To penalize polarized judgments, we down-weight the clarity score by the fraction of extreme votes:
    \begin{equation}
    S^{final} = \text{Clarity}_{clue} \times \Big(1 - \frac{\#\{s=1 \lor s=5\}}{n}\Big).
    \end{equation}
    This ensures that clues judged simultaneously ``too vague'' and ``too obvious'' receive low final scores even if their median appears moderate.
\end{enumerate}

The resulting $S^{final}$ value is averaged across all evaluated clues to yield the ``Clarity'' score reported in Table~\ref{tab:human_eval}.

\newpage
\section{DixitArena Environment Details}
\label{app:setup}

\paragraph{Data and Card Pool.}
We use a custom-made set of 84 Dixit-style illustrations (\texttt{1.png}–\texttt{84.png}), stored under \texttt{images/}. Each image is in PNG format (305$\times$460, RGBA, 8-bit per channel) and transmitted to the API via Base64 encoding without preprocessing. In each match, four players are dealt 4 cards each from a shuffled pool; in every round, the storyteller selects one target and the three guessers each contribute one distractor. These four images are shuffled using \texttt{random.shuffle()} to form the candidate set. Guessers select among them (excluding their own card), with distractor sampling guided by each model’s semantic matching rather than random choice. This ensures diversity and realism in distractor quality.

\paragraph{Game Flow.}
The game proceeds in rounds: (i) the storyteller selects a target and produces a clue, (ii) the other players submit distractors, (iii) the candidate set is shuffled and displayed, (iv) guessers select a card, and (v) scoring is applied. The scoring rules are: partial correct $\to$ storyteller +3 and correct guessers +3; all correct $\to$ storyteller 0, guessers +3; all wrong $\to$ storyteller 0, guessers +2; any distractor chosen $\to$ card owner +1. Each match lasts 24 rounds (two phases of 12). In Phase 2, players swap hands (P1$\leftrightarrow$P3, P2$\leftrightarrow$P4) to eliminate hand-quality bias. The evaluation spans 21 matches in a full round-robin, including self-play.

\paragraph{Randomness and Reproducibility.}
We fix the random seed at 42 for shuffling and candidate order. Single-run evaluation produces 504 rounds in total (21 matches × 24 rounds). All rounds, decisions, and scores are recorded in structured JSON, ensuring exact reproducibility given the same seed and configuration.

\paragraph{Fairness Guarantees.}
Fairness is achieved through hand-swap across phases, equal storyteller turns for all players, candidate shuffling to prevent position bias, and full round-robin coverage (including self-play) to eliminate matchup imbalance.


\paragraph{Open Source Resources.}
We plan to release code, configuration files, and sample logs for reproducibility. A minimal demo script (\texttt{simple\_demo.py}) reproduces a 2-round match for quick verification.

\newpage
\section{DixitBench Curation Details}
\label{app:benchmark}

DixitBench serves as an auxiliary benchmark that evaluates how well models can discriminate semantically similar yet subtly distinct visual descriptions under controlled difficulty levels.

\paragraph{Data Curation Pipeline.}
We first curate captions for all 84 Dixit-style illustrations using a vision-language model prompted to \textit{``describe this artwork implicitly with a single abstract phrase.''} Each caption represents a mid-level abstraction between a single-word concept and a full-sentence description, reflecting the ambiguous, poetic quality typical of Dixit gameplay.

\paragraph{Embedding and Similarity Computation.}
All captions are encoded into 384-dimensional semantic embeddings using the \texttt{sentence-transformers/all-MiniLM-L6-v2} model. Pairwise cosine similarities are computed to produce an $84 \times 84$ similarity matrix. This matrix serves as the foundation for controlled distractor sampling based on semantic closeness.

\paragraph{Difficulty-Based Distractor Sampling.}
For each target image $i$, we identify distractor candidates by ranking all other images according to cosine similarity:
\begin{itemize}
    \item \textbf{Hard:} top 5 most similar captions (ranks 1–5)
    \item \textbf{Easy:} randomly sampled from ranks 30–80
\end{itemize}
This design creates a clear contrast between high-confusability and low-confusability distractor sets. Each target thus yields two distractor groups, resulting in \textbf{168 benchmark items} (84 images $\times$ 2 difficulty levels).


\paragraph{Human Evaluation.}
To validate benchmark quality, we conducted human annotation on 20 sampled items (covering both difficulty levels). Annotators rated (i) caption-image coherence, (ii) distractor plausibility, and (iii) difficulty consistency. Average agreement exceeded 0.8 (Cohen’s $\kappa$), confirming the benchmark’s semantic validity.

This dataset complements DixitArena by isolating the perception component of abductive reasoning—testing whether a model can distinguish subtle conceptual differences in metaphorical phrasing before interacting in a multi-agent setting. The phrase-level dataset and scripts will be released alongside our main benchmark to facilitate reproducibility.

\newpage
\section{Qualitative Case Studies}
\label{app:case}

This appendix presents representative examples of actual game rounds from our DixitArena evaluation, illustrating both successful and failed storytelling strategies across different models.

\subsection{Case 1: Successful Storytelling (Partial-Correct)}
\textbf{Model:} Gemma3-27B (Storyteller) \\
\textbf{Round:} Match 3, Phase 1, Round 3 \\
\textbf{Clue:} ``A delicate hope, reaching for something unseen, supported by a fragile network.'' \\
\textbf{Target:} Image 35.png \\
\textbf{Outcome:} 2/3 listeners guessed correctly (Partial-Correct). \\
\textbf{Insight:} Balanced ambiguity with metaphorical language guided some listeners while misleading others.

\subsection{Case 2: Failed Storytelling (All-Correct, Too Obvious)}
\textbf{Model:} Qwen2.5-32B (Storyteller) \\
\textbf{Round:} Match 7, Phase 1, Round 5 \\
\textbf{Clue:} ``A knight in shining armor on horseback, holding a spear, as a tentacle rises from the pages of a book.'' \\
\textbf{Target:} Image 10.png \\
\textbf{Outcome:} 3/3 listeners guessed correctly (All-Correct). \\
\textbf{Insight:} Over-specific literal description eliminated ambiguity, leaving no room for alternative interpretations.

\subsection{Case 3: Successful Storytelling (Poetic Ambiguity)}
\textbf{Model:} Gemma3-27B (Storyteller) \\
\textbf{Round:} Match 7, Phase 1, Round 3 \\
\textbf{Clue:} ``A fleeting glimpse of a world unseen.'' \\
\textbf{Target:} Image 11.png \\
\textbf{Outcome:} 1/3 listeners guessed correctly (Partial-Correct). \\
\textbf{Insight:} Poetic abstraction created interpretive space; one aligned listener succeeded while others were misled.

\subsection{Case 4: Failed Storytelling (All-Wrong, Too Vague)}
\textbf{Model:} Gemma3-12B (Storyteller) \\
\textbf{Round:} Match 6, Phase 1, Round 3 \\
\textbf{Clue:} ``A forgotten trophy, watching over a curious secret.'' \\
\textbf{Target:} Image 56.png \\
\textbf{Outcome:} 0/3 listeners guessed correctly (All-Wrong). \\
\textbf{Insight:} Semantic mismatch: the clue did not align with the target image, and even the same-model listener failed.
\subsection{Overall Insights}
Across these cases, we observe distinct success and failure patterns. Successful clues (Cases 1 and 3) typically rely on metaphorical or poetic language that balances clarity and ambiguity, allowing some listeners to be guided while others are misled. In contrast, failures arise from two opposite directions: over-specification (Case 2), where excessive literal detail leaves no room for interpretation, and semantic mismatch (Case 4), where vague or misaligned clues fail to connect with the intended target. These qualitative examples complement the aggregate statistics by revealing \textit{why} models succeed or fail in abductive storytelling.

\newpage
\section{Additional Results}

\subsection{Evaluation under Entailment Scoring}
\label{app:entailment}
To further assess the robustness of listener evaluation, 
we additionally tested models on DixitBench using an alternative 
\emph{entailment scoring} scheme (Table~\ref{tab:caption_similarity_results_entailment}). 
Unlike the standard \emph{direct selection} setting where models must 
choose a single best-matching image, the entailment variant requires 
assigning independent plausibility scores (0--100) to all candidates. 
This approach evaluates a model’s ability to produce 
well-calibrated absolute judgments rather than comparative preferences.

Overall, results show that entailment scoring yields moderately lower accuracies across all models—particularly on the \emph{Hard} subset—while preserving the same relative ranking trends observed under direct selection.
GPT-4o and Gemini-2.5-Flash remain the strongest performers, suggesting that although VLMs handle relative choice well, 
their absolute entailment calibration remains underdeveloped.
This supports our earlier finding that current VLMs are optimized for \textit{comparative abductive reasoning} rather than calibrated entailment verification.

\begin{table}[h]
\centering
\small
\setlength{\tabcolsep}{10pt}
\begin{tabular}{lccc}
\toprule
\textbf{Model} & \textbf{Easy} & \textbf{Hard} & \textbf{Total} \\
\midrule
Qwen2.5-VL-7B     & 22.47 & 18.19 & 20.33 \\
Qwen2.5-VL-32B    & 58.37 & 49.81 & 54.09 \\
Gemma3-12B        & 56.03 & 40.51 & 48.27 \\
Gemma3-27B        & 58.23 & 50.41 & 54.32 \\
Gemini-2.5-Flash  & \underline{61.21} & \underline{53.43} & \underline{57.32} \\
GPT-4o            & \textbf{66.69} & \textbf{52.41} & \textbf{59.55} \\
\bottomrule
\end{tabular}
\caption{Performance on \textit{DixitBench} under the \emph{entailment scoring} scheme. 
Scores are moderately lower than under direct selection, 
especially on the \emph{Hard} subset, while relative ranking remains consistent.}
\label{tab:caption_similarity_results_entailment}
\end{table}

\newpage
\subsection{Leave-One-Listener-Out Results}
\label{app:lolo}
Table~\ref{tab:lolo} reports the leave-one-listener-out (LOLO) analysis. 
For each model, we compare the original storyteller score with the average score 
when one listener is removed at a time. 
The small average differences and high stability indices confirm 
that no single listener dominates the outcome.

\begin{table}[h]
\centering
\scriptsize
\begin{tabular}{lccccc}
\toprule
\textbf{Model} & \textbf{Orig. Score} & \textbf{Avg} $\Delta$ & \textbf{Std} $\Delta$ & \textbf{Stability} \\
\midrule
Qwen2.5-VL-7B    & 21 & -0.08 & 0.28 & 0.908 \\
Qwen2.5-VL-32B   & 27 & -0.11 & 0.31 & 0.897 \\
Gemma3-12B       & 60 & -0.24 & 0.43 & 0.858 \\
Gemma3-27B       & 72 & -0.29 & 0.45 & 0.849 \\
Gemini-2.5-Flash & 69 & -0.27 & 0.45 & 0.851 \\
GPT-4o           & 51 & -0.20 & 0.40 & 0.866 \\
\bottomrule
\end{tabular}
\caption{Leave-one-listener-out (LOLO) storyteller scores. 
Stability is defined as $1 - \text{Std}(\Delta)/3.0$, ranging from 0--1 (higher = more robust).}
\label{tab:lolo}
\end{table}

    \end{document}